\documentclass{article}
\usepackage{ijcai25}

\usepackage{times}
\usepackage{soul}
\usepackage{url}
\usepackage[hidelinks]{hyperref}
\usepackage[utf8]{inputenc}
\usepackage[small]{caption}
\usepackage{graphicx}
\usepackage{amsmath}
\usepackage{amsthm}
\usepackage{booktabs}
\usepackage{algorithm}
\usepackage{algorithmic}
\usepackage[switch]{lineno}

\usepackage{multirow} % For multirow cells
\usepackage{url}            % simple URL typesetting
\usepackage{booktabs}       % professional-quality tables
\usepackage{amsfonts}       % blackboard math symbols
\usepackage{nicefrac}       % compact symbols for 1/2, etc.
\usepackage{microtype}      % microtypography
\usepackage{enumitem} 
\usepackage{subcaption}
\usepackage{amsmath}
\usepackage{cleveref}
\usepackage{makecell}
\crefname{figure}{Figure}{Figures}
\crefname{table}{Table}{Tables}
\usepackage[table]{xcolor}
\usepackage{csquotes}
\title{BMIP: Bi-directional Modality Interaction Prompt Learning for VLM}

\author{
Song-Lin Lv$^{1,3}$ \and
Yu-Yang Chen$^{1,3}$\and
Zhi Zhou$^3$\and
Ming Yang$^{2,3}$ \And Lan-Zhe Guo\thanks{Corresponding author}$^{,1,3}$
\affiliations
$^1$School of Intelligence Science and Technology, Nanjing University, China. \\
$^2$School of Artificial Intelligence, Nanjing University, China. \\
$^3$National Key Laboratory for Novel Software Technology, Nanjing University, China.
\emails
\{lvsl,chenyy,zhouz,yangm,guolz\}@lamda.nju.edu.cn
}

\begin{document}

\maketitle

\begin{abstract}
Vision-language models (VLMs) have exhibited remarkable generalization capabilities, and prompt learning for VLMs has attracted great attention for the ability to adapt pre-trained VLMs to specific downstream tasks. However, existing studies mainly focus on single-modal prompts or uni-directional modality interaction, overlooking the powerful alignment effects resulting from the interaction between the vision and language modalities. To this end, we propose a novel prompt learning method called \underline{\textbf{B}}i-directional \underline{\textbf{M}}odality \underline{\textbf{I}}nteraction \underline{\textbf{P}}rompt (BMIP), which dynamically weights bi-modal information through learning the information of the attention layer, enhancing trainability and inter-modal consistency compared to simple information aggregation methods. To evaluate the effectiveness of prompt learning methods, we propose a more realistic evaluation paradigm called \emph{open-world generalization} complementing the widely adopted \emph{cross-dataset transfer} and \emph{domain generalization} tasks. Comprehensive experiments on various datasets reveal that BMIP not only outperforms current state-of-the-art methods across all three evaluation paradigms but is also flexible enough to be combined with other prompt-based methods for consistent performance enhancement.
\end{abstract}

\section{Introduction}
Vision-language models, such as CLIP~\cite{CLIP}, pre-trained on vast web-scale text-image datasets, have demonstrated impressive zero-shot capabilities in a variety of downstream image classification tasks \cite{jia2021scaling,CLIP}. Concurrently, various prompt learning methods for VLMs have been proposed to further improve the performance of VLMs on specific tasks by exploiting a small amount of labeled data for the target task to finetune learnable prompts while keeping other parameters frozen~\cite{jia2022visual,khattak2023maple,zhou2022conditional,zhou2022learning}. 

\begin{figure*}[t]
  \centering
  \begin{subfigure}{.4\textwidth}
    \centering
    \includegraphics[width=\linewidth]{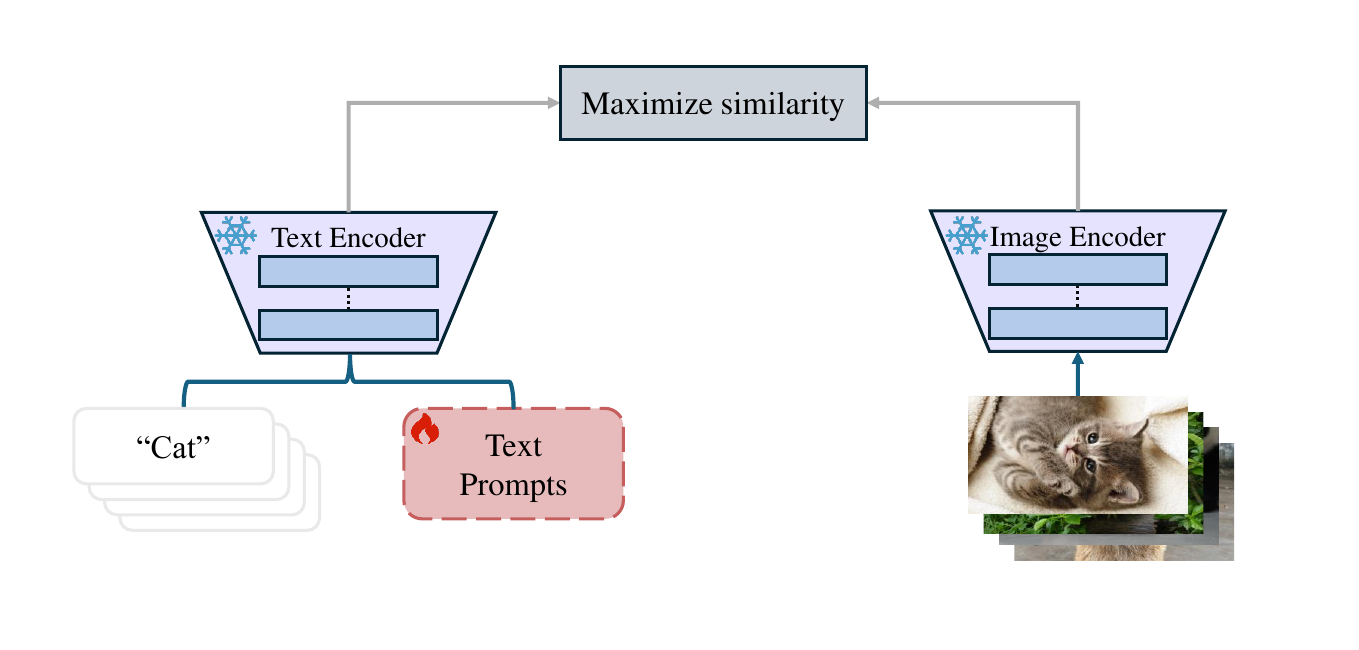}
    \caption{Language Prompt}
    \label{fig:sub1}
  \end{subfigure}%
  \begin{subfigure}{.4\textwidth}
    \centering
    \includegraphics[width=\linewidth]{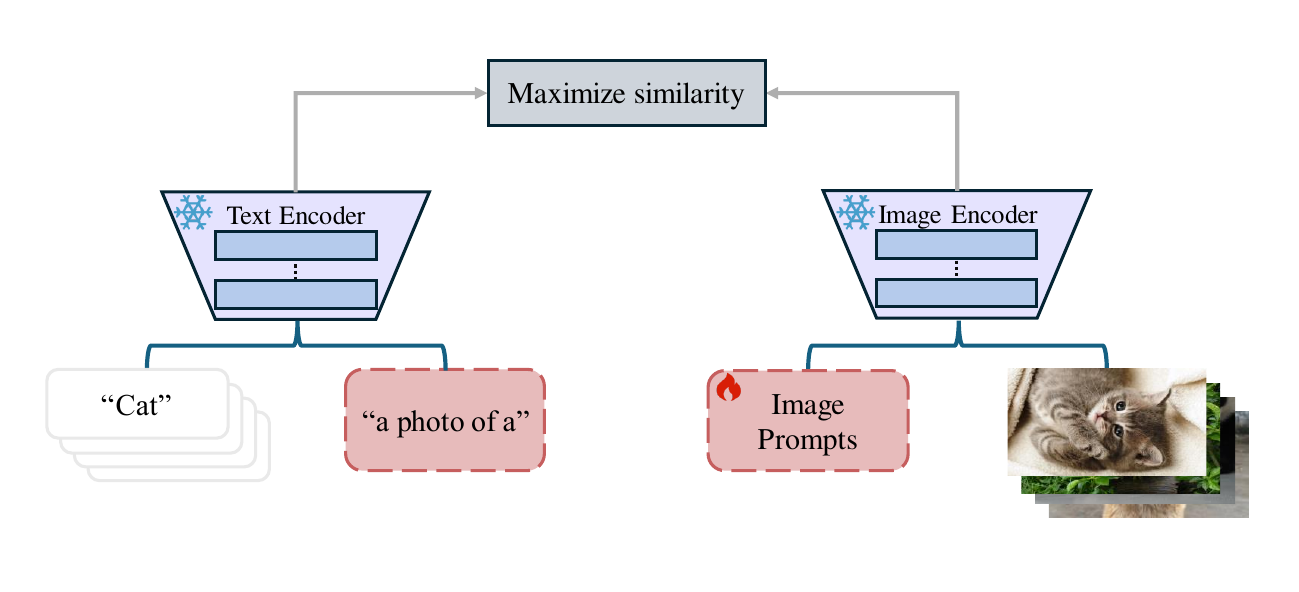}
    \caption{Vision Prompt}
    \label{fig:sub2}
  \end{subfigure}
  \begin{subfigure}{.4\textwidth}
    \centering
    \includegraphics[width=\linewidth]{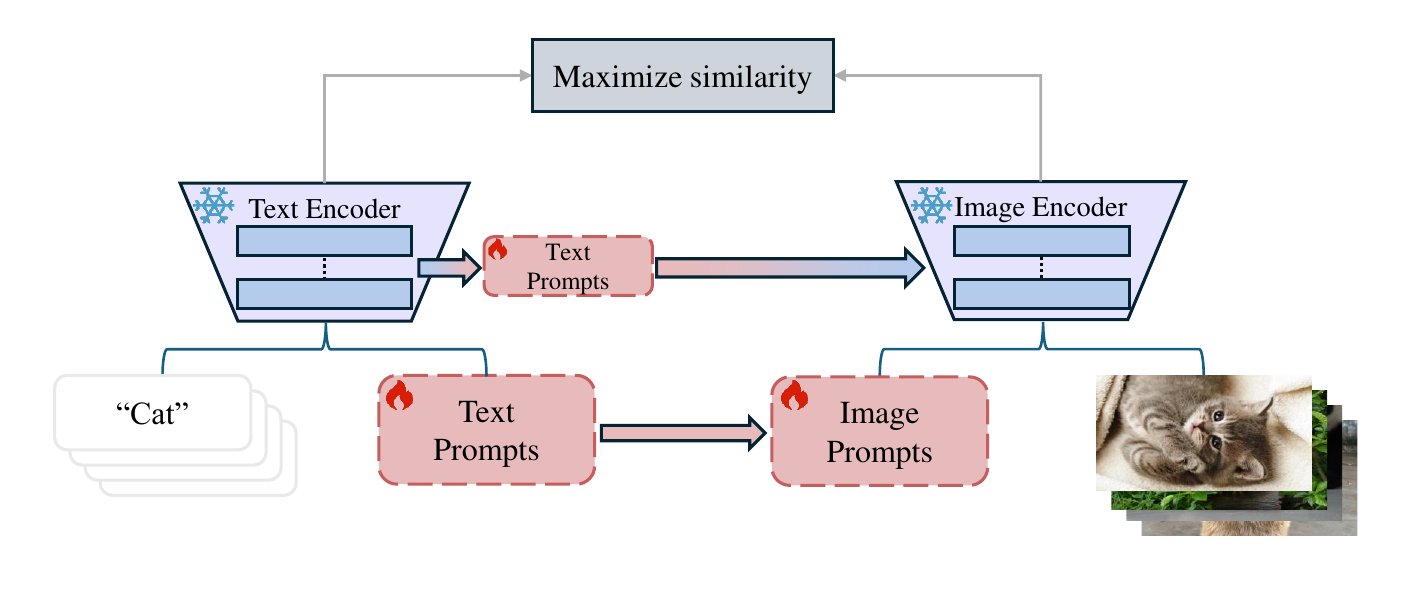}
    \caption{Uni-direction Modality Interaction Prompt}
    \label{fig:sub3}
  \end{subfigure}%
  \begin{subfigure}{.4\textwidth}
    \centering
    \includegraphics[width=\linewidth]{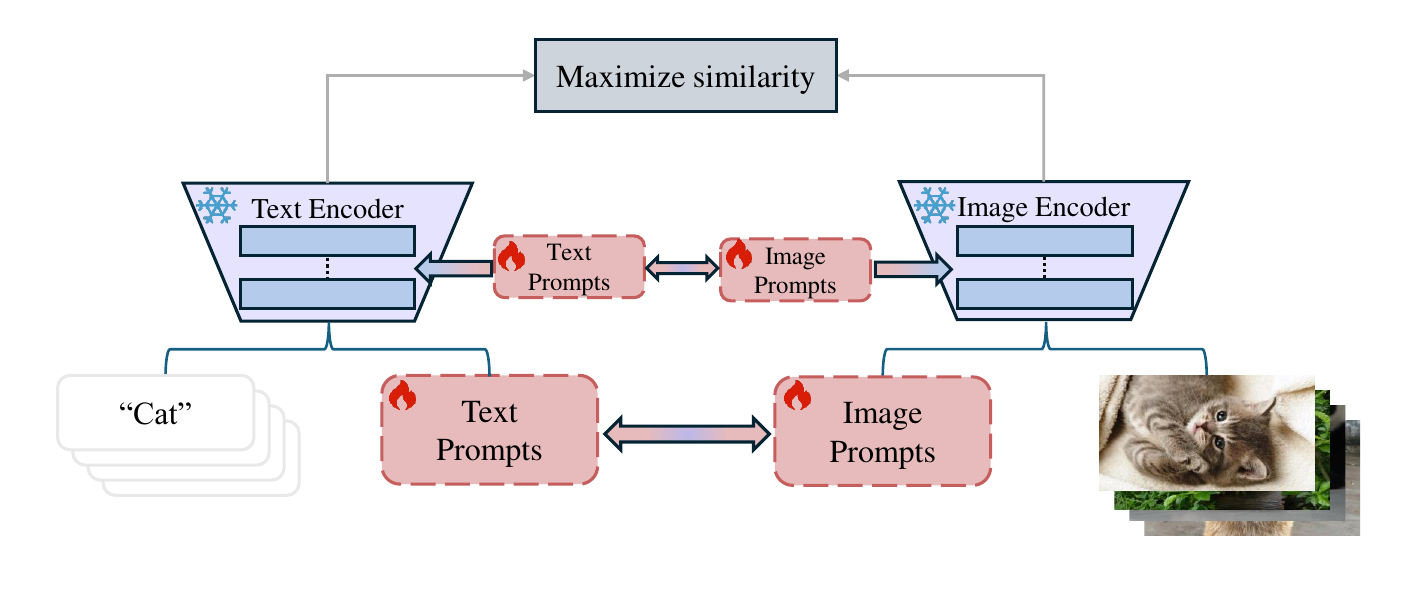}
    \caption{Bi-direction Modality Interaction Prompt}
    \label{fig:sub4}
  \end{subfigure}
  \caption{(a) Adapting language modality representations for downstream tasks through language prompt learning. (b) Adapting vision modality representations for downstream tasks through vision prompt learning. (c) Achieving uni-directional modality interaction by converting language prompts into vision prompts. (d) Achieving bi-directional modality interaction by aligning two modalities' representations through aggregating information between the vision and language modalities. (\includegraphics[width=0.015\textwidth]{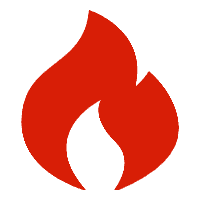}: learnable;~\includegraphics[width=0.015\textwidth]{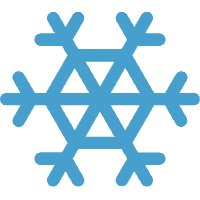}: frozen during training)}
  \label{fig:whole}
\end{figure*}

Previous methods mainly learn prompts on a single modality, either language (\cref{fig:whole}:\subref{fig:sub1}) or vision (\cref{fig:whole}:\subref{fig:sub2}), overlooking the interaction between the two modalities, which is crucial for preserving alignment between the vision and language. Many studies~\cite{zang2022unified,varpro,lee2023read} have reported that language prompt learning underperforms on datasets with high intra-class visual variances, while vision prompt learning struggles on datasets with small inter-class textual variances. For example, the EuroSAT dataset exhibits high intra-class visual variances, which is challenging for text-only methods to handle effectively. There are also some studies~\cite{rao2022denseclip,kan2023knowledge,khattak2023maple} that attempt to transfer prompts from language to vision modality (\cref{fig:whole}:\subref{fig:sub3}) to exploit modality interaction. However, these methods only achieve alignment provided by language prompts alone and fail to fully consider the effect of the interaction between the two modalities. To the best of our knowledge, how to exploit the bi-directional interaction between the two modalities in VLMs remains a significant yet unsolved challenge.

To address the lack of focus on multi-modal consistency in single-modal interaction methods, we propose a novel approach called \underline{\textbf{B}}i-directional \underline{\textbf{M}}odality \underline{\textbf{I}}nteraction \underline{\textbf{P}}rompt (BMIP), as illustrated in Figure~\ref{fig:whole}:\subref{fig:sub4}. Given the absence of existing studies on multi-modal interaction in the VLM field and the limitations of simple aggregation functions—such as low information utilization and potential information distortion—we designed an innovative aggregation function for multi-modal interaction. This function leverages the relationship between the model’s attention layer outputs and prompt importance, dynamically balancing dual-modal information through adaptive weighting to solve the above difficulties.

To evaluate the effectiveness of prompt learning methods in VLMs, inspired by open-world settings~\cite{opsetting,knowkd}, we propose a novel evaluation paradigm called \emph{open-world generalization}. Different from the previous \emph{base-to-new class generalization} task~\cite{zhou2022learning,khattak2023maple,yao2023visual}, which evaluates base and new classes separately, the new paradigm does not pre-determine whether the data belongs to base or new classes, leading to a more realistic evaluation.
Experimental results on 15 benchmarks demonstrated that the BMIP architecture achieves significant performance improvement compared to state-of-the-art (SOTA) prompt learning methods, especially in \emph{open-world generalization} evaluation paradigm. Specifically, BMIP addresses the poor performance of single-modal prompt learning methods when dealing with unbalanced datasets of text and images, such as EuroSAT~\cite{helber2019eurosat}
and Flowers102~\cite{nilsback2008automated}, which is in line with the motivation. 
Additionally, as an enhanced algorithm for MaPLe, BMIP can serve as a foundational framework that integrates with any other prompt learning methods to further improve their performance.

In summary, the main contributions of this work include:
\begin{itemize}[itemsep=1pt, parsep=1pt]
\item We analyze the limitations of previous prompt learning methods and propose a novel technique that leverages bi-directional modality interaction, which enhances the alignment between vision and language modalities and paves the way for further exploration of information aggregation in other multi-modal models.

\item To evaluate prompt learning methods, we propose a more realistic evaluation paradigm called \emph{open-world generalization}. We believe this could facilitate more realistic evaluations of prompt learning methods and promote related research. 

\item We conduct comprehensive experiments on 15 benchmarks. The results demonstrated that BMIP achieves SOTA performance across all tasks, and is flexible enough to be combined with other prompt learning methods, consistently enhancing their performance.
\end{itemize}

\section{Related Works}
\textbf{Vision-Language Models.} VLMs~\cite{jia2021scaling,CLIP,Flamingo,filip} have demonstrated outstanding performance on a wide range of image classification tasks under zero-shot and few-shot settings. 
% By utilizing a substantial amount of text-image data, these models leverage the common embedding space of text and image representations to achieve cross-modal alignment effectively. 
% For example, CLIP and ALIGN respectively use \texttildelow400M and \textasciitilde1B image-text pairs to train multi-modal networks. 
While these pre-trained VLMs have learned generalized representations of images and texts, adapting them quickly and effectively to downstream tasks remains a challenging problem. 
Existing approaches to exploring finetuning VLMs to downstream tasks using a small number of parameters and data can be categorized into two types: adapter tuning~\cite{zhang2021tip,graphadapter,taskresidual} and prompt learning ~\cite{shu2022test,zhou2022learning,zhou2022conditional} which is the focus of our work. 

\textbf{Prompt Learning in Vision Language Models.} Due to the large parameter size of VLMs and the limited availability of training data for downstream tasks, it is impractical to finetune all parameters of the VLMs to adapt them to these tasks. Inspired by the success of prompt learning in NLP~\cite{he2022hyperprompt,li2021prefix}, many researchers have proposed to adapt VLMs by learning the prompts in end-to-end training. As the pioneering work, CoOp~\cite{zhou2022conditional} for the first time introduces the learnable prompt to transfer the task-specific knowledge to VLMs. To improve the generalization of the learnable language prompt in CoOp, CoCoOp~\cite{zhou2022learning} and VPT~\cite{jia2022visual} generate a vision-conditional prompt by fusing the image feature and the learnable language prompts. 
% Then UPT~\cite{zang2022unified} proposes a unified prompt learning strategy for text and image encoders. 
kgCoOp~\cite{yao2023visual}, ProGrad~\cite{zhu2023prompt} and other methods~\cite{lee2023read,tan2024compound} are another prompt-based methods for VLMs. MaPLe~\cite{khattak2023maple}, conducts language-vision prompt learning by jointly applying prompt learning to both the vision and text encoders, simultaneously refining the text and image representations for adapting to downstream tasks. Our proposed method focuses on addressing the imbalance in alignment between the vision and language modalities caused by the lack of interaction in existing algorithms.

\section{Preliminary}
Our approach BMIP is proposed based on CLIP, learning both language and vision prompts. Therefore, before introducing the proposal, we revisit the main ideas of CLIP, language prompt learning, and vision prompt learning.

\textbf{CLIP.} CLIP is developed to align visual and textual data in a common embedding space. CLIP consists of two encoders: an image encoder denoted as \(f\) and a text encoder denoted as \(g\). During the training phase, the encoders extract feature representations \(f(I)\) and \(g(E_w(T))\) from an input image \(I\) and its corresponding text caption \(T\), respectively. The term \(E_w\) represents the word embedding layer, tasked with transforming words into vector representations.

During the zero-shot classification phase, CLIP begins with an image \(I\) and a set of hand-designed text captions \([T_1, T_2,\dots, T_N]\), formatted as \enquote{\(a\ photo\ of\ a\ [CLASS_i]\)}, where \enquote{\(a\ photo\ of\ a\)} is a hand-designed template and \([CLASS_i]\) specifies a class from \(N\) candidate image categories. The image and captions are processed by their respective encoders to extract features, allowing for the computation of class prediction probabilities as follows:
\begin{equation}
p(y=i|I)=\frac{\exp(cos(f(I), g(E_w(T_i)))/\tau)}{\sum_{j=1}^N
\exp(cos(f(I), g(E_w(T_j)))/ \tau)}  
\end{equation}
In this context, \(\tau\) is the temperature coefficient, and \(cos(\cdot,\cdot)\) represents the cosine similarity between features.

\textbf{Language Prompt Learning.} To effectively adapt VLMs to downstream tasks, language prompt learning aims to generate more adaptive classifiers, without the need to finetune the text encoder \(g\). For instance, some studies \cite{lu2022prompt,zhou2022conditional,zhou2022learning} employ learnable prompts \(P = [t_1,t_2,\dots,t_b]\) to replace hand-designed language prompt templates, where \(t\) represents the prompt vector, and \(b\) specifies the prompt's length. Let \([c_1, c_2, \dots, c_N]\) represent the word embeddings of class names. The corresponding prediction probability is calculated as follows:
\begin{equation}
p(y=i|I)=\frac{\exp(cos(f(I), g([P,c_i]))/\tau)}{\sum_{j=1}^N
\exp(cos(f(I), g([P,c_j]))/ \tau)}
\end{equation}
In this context, \([\cdot,\cdot]\) denotes the operation of concatenation. For each downstream task, the learnable prompt \(P\) is optimized via cross-entropy classification loss during the few-shot learning phase. As a result, updating the language prompt \(P\) will adjust the decision boundaries accordingly, utilizing the generated classifier for the downstream tasks.

\textbf{Vision Prompt Learning.} Analogous to the language modality, vision prompt learning incorporates vision prompt vectors \(\tilde{P}\) to extract more representative visual features. To associate images with these vectors, the image embedding layer transforms image patches \(I = [I_1, I_2,\dots, I_{m}]\) into image patch embeddings \(E = [e_1,e_2,\dots,e_{m}]\), with \(m\) represents the number of image patches. The output from the \(i^{th}\) layer of image encoder \(f_i\) is expressed by the equation:
\begin{equation}
\left[ CLS_i,E_i,\tilde{P}_i \right] \ =\ f_i\left( \left[ CLS_{i-1},  E_{i-1},\tilde{P}_{i-1} \right] \right) 
\end{equation}
Here, \(CLS\) represents the image class token, which will be projected by a projection head to obtain the image feature. 
\begin{figure*}[t]
\centering
\includegraphics[width=.8\textwidth]{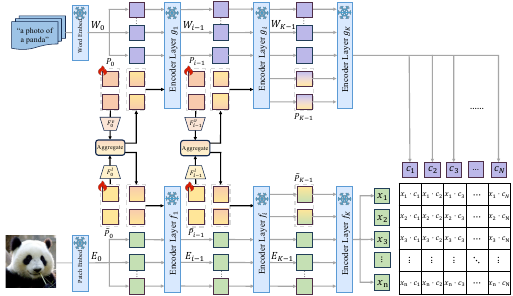}
\caption{Overview of proposed BMIP method. BMIP finetunes the layered prompts for the vision and language branches while freezing the rest of the model parameters. It utilizes an aggregation function to modulate cross-modal prompt influences, thus enabling effective information exchange between the two modalities.}
\label{figure2}
\end{figure*}

\section{Bi-directional Modality Interaction Prompt}
Many studies~\cite{varpro,lee2023read,zang2022unified} have reported that language prompt learning underperforms on datasets with high intra-class visual variances while vision prompt learning struggles on datasets with small inter-class textual variances. Recognizing the substantial influence of the multi-modal interaction, we propose the BMIP method from three intuitions: (1) Independent vision and language prompts facilitate the collection of information from their respective modalities, while projected prompts are substitutable for original prompts; (2) prompts of suitable depth can expand the scope of prompt information and curtail overfitting; and (3) effective interaction among modalities will mitigate the drawbacks of imbalanced single-modal information and promote better alignment between vision and language modalities. Therefore, the proposed BMIP is composed of three key components: deep language prompt learning, deep vision prompt learning, and vision language modality interaction. Figure \ref{figure2} illustrates the overall architecture of the BMIP framework, and we describe the details in the following. In the end, we analyze the rationale behind BMIP’s modality interaction.

\subsection{Deep Language Prompt Learning}
In contrast to traditional language prompt learning, deep language prompt learning introduces layered prompts to expand the scope of prompt information. Specifically, we introduce layered language prompts \(\{P_i \in \mathbb{R}^{1 \times b \times d_l}\}_{i=0}^J\), where \(J\), \(b\), and \(d_l\) indicates the depth, length, and dimension of the language prompts, respectively. The input to the initial layer assumes the structure \([P_0, W_0]\), where \(\{W_0 \in \mathbb{R}^{N \times x \times d_l}\}\) signifies the word embedding of the text \(T\). Here, \(x\) represents the number of words in a single caption from \(T\), with \(N\) indicating the total count of image categories. To provide comprehensive guidance on the word embeddings \(W\), we deploy language prompts to supersede prompts from prior layers in the first \(J\) layers of the text encoder. The inputs and outputs at the \(i^{th}\) layer of the text encoder \(g\) are as follows:
% The prompt \(P_0\) is replicated \(N\) times to align with \(W_0\), with \(N\) indicating the total count of image categories.
\begin{equation}
\left[ \_,W_i \right] =g_i\left( \left[ P_{i-1}, W_{i-1} \right] \right) \quad i=1,2,\dots,J    
\end{equation}
Beyond the \({J}^{th}\) layer, to prevent the model from excessive reliance on learnable prompts and overfitting, we use the prompts from the output of the preceding layer as the input to the next layer. The class feature \(z\) is obtained by projecting the class representation \([c_1, c_2, \dots, c_N]\) corresponding to the last output of the text encoder \(W_K\) to a common embedding space via the text projection head \(\mathrm{TextProj}\).
\begin{equation}
\left[ P_j,W_j \right] =g_j\left( \left[ P_{j-1}, W_{j-1} \right] \right) \quad j=J+1,\dots,K
\end{equation}
\begin{equation}
z=\mathrm{TextProj}\left([c_1, c_2, \dots, c_N]\right) 
\end{equation}
\subsection{Deep Vision Prompt Learning}
To gather information about the dataset and interact with the language modality, we introduce independent visual prompts. Deep vision prompt learning employs a set of vision prompts \(\{\tilde{P}_i \in \mathbb{R}^{1 \times b \times d_v}\}_{i=0}^J\), which match the depth and length of language branches, differing solely in the dimension of the vision prompts \(d_v\). In the first \(J^{th}\) layers of the image encoder \(f\), we use a learnable vision prompt to replace the output of the previous layer. These prompts are then concatenated with the class token \(CLS\) and the image patch embeddings \(E\), forming the input for the next layer, as shown below:
\begin{equation}
\begin{aligned}
\left[ CLS_i, E_i, \_ \right] &= f_i \left( \left[ CLS_{i-1}, E_{i-1}, \tilde{P}_{i-1} \right] \right) \\
i &= 1, 2, \dots, J
\end{aligned}
\end{equation}
After the \(J^{th}\) layer, similar to the language branch, the ensuing layer's input is the immediate output of its predecessor. Upon obtaining the ultimate class token \(CLS_K\), the image projection head, denoted as \(\mathrm{ImageProj}\), is employed to map the final image feature \(x\) to the common embedding space. The formula from the \((J+1)^{th}\) layer is as follows:
\begin{equation}
\begin{aligned}
\left[ CLS_j,E_j,\tilde{P}_{j} \right] &= f_j\left( \left[ CLS_{j-1},E_{j-1},\tilde{P}_{j-1} \right] \right) \\ 
j &= J+1,\dots,K
\end{aligned}
\end{equation}
\begin{equation}
x=\mathrm{ImageProj}\left( CLS_K \right) 
\end{equation}

\subsection{Vision Language Modality Interaction}
As a novel multimodal interaction approach to prompt learning, it is essential for BMIP to design with an interactive architecture that promotes effective information aggregation. Under this objective, this architecture consists of three key components: a language projection head \(F_l\), a vision projection head \(F_v\), and an aggregation function.

The most straightforward idea for aggregating information might include adding the aggregated information to the original prompts using unit weights or attention weights or simply connecting this information. Unfortunately, these methods driven by linear addition and modifications either fail to allocate the importance of the two modality inputs effectively or cannot easily highlight the internal impact of the prompts, making efficient information aggregation unattainable for effective information aggregation from both modalities.

To solve this problem, we propose a learnable aggregation function using the output weight of the attention layer for effective information aggregation from both modalities. This aggregation function uses learnable modules to dynamically generate weights for different modality prompts, giving greater emphasis to the prompts that the attention layer focuses on. Specifically, the vision and language projection heads produce transformed vision and language information, represented as \(\{F_v(P_{i}), F_l(\tilde{P}_{i})\}\). The vision and language attention weight (\(w_v\), \(w_l\)) are extracted from the output of the current attention layer, representing the degree of attention given by other inputs to the current prompt. Subsequently, we trained modality-specific \(1 \times 1\) linear layers, \(L_l\) and \(L_v\), to learn the relationship between attention weights and substitution weights (\(w_l, w_v\)), ultimately generating unique dynamic weights for each prompt. The formal expression is as follows:

\begin{equation}
\begin{array}{c}
w_v = L_v(A_v), \\
\tilde{P}_{i}' = w_v * \tilde{P}_{i} + (1 - w_v) * F_v(P_{i}).
\end{array}
\end{equation}

\begin{equation}
\begin{array}{c}
w_l = L_l(A_l), \\
P_i' =\left[w_l * P_i + (1 - w_l) * F_l(\tilde{P}_i)\right].
\end{array}
\end{equation}

where \(\{P_{i}'|_{i=1}^{J}\}\) and \(\{\tilde{P}_{i}'|_{i=1}^J\}\) represent augmented deep language and vision prompts. After implementing the aggregation function, the first \(J\) layers of BMIP process inputs and outputs for the language and vision modalities as described subsequently:
\begin{equation}
\begin{aligned}
\left[ \_,W_i \right] &= g_i\left( \left[P_{i}', W_{i-1} \right] \right) 
\end{aligned}
\end{equation}
\begin{equation}
\left[ CLS_i,E_i,\_ \right] =f_i\left( \left[ CLS_{i-1},E_{i-1},\tilde{P}_{i}' \right] \right)
\end{equation}
Beyond the \(J^{th}\) layer, subsequent layers use the output of the predecessor as input to obtain the final prediction. Using the proposed aggregation function, vision and language encoders can receive weighted prompt input, facilitating alignment of vision and language. Additionally, the learnable linear layers eliminate the need for manually tuning prompt weights, ensuring high reliability and greater flexibility. Furthermore, BMIP can enhance existing prompt fine-tuning methods by serving as a more powerful foundational model than MaPLe, leveraging its effectively balanced multimodal prompts.

\subsection{Analysis}
CoCoOp and subsequent prompt learning methods demonstrate that for well-aligned pre-trained VLMs, fine-tuned prompts are interchangeable. For instance, CoCoOp incorporates image information into text prompts, while MaPLe uses transformed language prompts as vision prompts, which reflects that vision and language prompts can be simultaneously optimized and mutually substituted, as demonstrated in the Ablation Studies.

In practice, especially when approaching the convergence point, the magnitude of \(\frac{\partial L}{\partial w}\) is usually very close to zero, indicating that the probability of staying around \(w\) is large. In other words, when the attention weight is equal to zero, this prompt will almost become redundant during the later training process. Therefore, using \(w\) to combine the original prompt with aggregated information will enhance the trainability of the model and facilitate modal alignment. We immediately have the following corollary,

\textbf{Corollary 1} \textit{If the minimal of weights implies \(w=0\), then the prompt combining will only decrease the training loss, i.e. \(min_{P'}L \leq min_{P}L\), given the sufficiently expressive \(min_{P'}L\) and \(min_{P}L\) which denote the cases with and without prompt combining, respectively.}

These analysis suggests that replacement prompts enhance BMIP's trainability, maintaining image text alignment while adapting to downstream tasks.

\section{Experiments}

\subsection{Experimental Setup}
\textbf{Evaluation Paradigm.} There are three widely adopted tasks to evaluate prompt learning methods: \emph{generalization from base to novel classes, cross-dataset transfer, and domain generalization}~\cite{zhou2022conditional,khattak2023maple,lee2023read,yao2023visual}.
Generalization from base to novel classes task evaluates the performance of models on base and new classes separately. Although this evaluation paradigm can comprehensively evaluate the performance of both base and new classes, it lacks practicality for real-world applications since downstream tasks cannot determine whether the data belongs to base or new classes in advance, which is a major metric of a model's ability to generalize. 
Therefore, we propose a more realistic evaluation paradigm, termed \emph{open-world generalization}, which introduces a metric used in open-world settings~\cite{opsetting,knowkd}, the simultaneous evaluation of base classes and new classes, to the generalization from base to novel classes task. This new paradigm assesses the model’s performance on an unknown distribution composed of both base and novel classes.
The \emph{cross-dataset transfer} task demonstrates the model's zero-shot generalization to novel datasets, while the \emph{domain generalization} task assesses the model’s robustness to out-of-distribution data. 

\textbf{Dataset.} We assessed 11 image recognition datasets in the open-world generalization and cross-dataset transfer tasks, encompassing ImageNet (\citeyear{deng2009imagenet}), Caltech101 (\citeyear{fei2004learning}), OxfordPets (\citeyear{parkhi2012cats}), StanfordCars (\citeyear{krause20133d}), Flowers102 (\citeyear{nilsback2008automated}), Food101 (\citeyear{bossard2014food}), FGVC-Aircraft (\citeyear{maji2013fine}), SUN397 (\citeyear{xiao2010sun}), UCF101 (\citeyear{soomro2012ucf101}), DTD (\citeyear{cimpoi2014describing}), and EuroSAT (\citeyear{helber2019eurosat}). For domain generalization task, ImageNet serves as our source dataset, while 4 variants ImageNetV2 (\citeyear{recht2019imagenet}), ImageNet-Sketch (\citeyear{imagenetsketch}), ImageNet-A (\citeyear{hendrycks2021natural}), and ImageNet-R (\citeyear{hendrycks2021many}) serve as the target datasets.

\textbf{Compared Methods.} There are several methods similar to BMIP that use prompts exclusively to finetune VLMs, such as CoOp, PLOT, ProDA, and MaPLe; therefore, we focus on benchmarking against four current representative methods: CLIP, CoOp, CoCoOp, and MaPLe, the SOTA method. It is worth noting that BMIP is orthogonal to studies that employ regularization techniques and can be integrated with these methods to enhance prompt learning optimization. We report the performance improvements brought by BMIP in ablation studies. The \textbf{implementation details} are in Appendix A.

\begin{table*}[t]
    \centering
    \setlength{\tabcolsep}{0.6mm}
    \fontsize{10}{12}\selectfont
    \begin{tabular}{l|cc|ll|ll|ll}
    \toprule
    & \multicolumn{2}{c|}{Average}  & \multicolumn{2}{c|}{ImageNet}  & \multicolumn{2}{c|}{Caltech101} & \multicolumn{2}{c}{OxfordPets}  \\
    & \multicolumn{1}{c}{HM}      & \multicolumn{1}{c|}{Acc.}  & \multicolumn{1}{c}{HM}      & \multicolumn{1}{c|}{Acc.}  & \multicolumn{1}{c}{HM}      & \multicolumn{1}{c|}{Acc.}  & \multicolumn{1}{c}{HM}      & \multicolumn{1}{c}{Acc.}   \\ \midrule 
    CLIP & 70.84  & 63.92 & 70.20 $\pm$ 0.00 & 66.73 $\pm$ 0.00 & 95.41 $\pm$ 0.00 & 92.90 $\pm$ 0.00 & 92.93 $\pm$ 0.00 & 88.03 $\pm$ 0.00\\ 
    CoOp & 72.14 & 65.57 & 64.95 $\pm$ 1.11 & 61.79 $\pm$ 1.09 & 95.96 $\pm$ 0.39 & 93.24 $\pm$ 0.68 & 95.38 $\pm$ 0.33 & 89.61 $\pm$ 0.34\\ 
    CoCoOp & 74.72 & 67.67 & 72.71 $\pm$ 0.33 & 69.41 $\pm$ 0.36 & 95.55 $\pm$ 0.24 & 93.43 $\pm$ 0.37 & 95.71 $\pm$ 0.76  & 90.24 $\pm$ 1.32 \\ 
    MaPLe & 78.22 & 71.76 & 73.60 $\pm$ 0.12 & 70.30 $\pm$ 0.16          
    & 96.47$\pm$ 0.41 & 94.67 $\pm$ 0.33         
    & \textbf{96.68 $\pm$ 0.12}   & \textbf{92.60 $\pm$ 0.49}          \\ \rowcolor{gray!20}
    BMIP  & \textbf{79.04} & \textbf{72.17}             
    & \textbf{73.47 $\pm$ 0.11}  & \textbf{70.23 $\pm$ 0.06}  
    & \textbf{96.54 $\pm$ 0.25}  & \textbf{94.73 $\pm$ 0.23}
    & 96.40 $\pm$ 0.19 & 92.53 $\pm$ 0.32 \\ 
    \midrule
    \midrule
    & \multicolumn{2}{c|}{StandfordCars} & \multicolumn{2}{c|}{Flowers102} & \multicolumn{2}{c|}{Food101} & \multicolumn{2}{c}{FGVCAircraft} \\
    & \multicolumn{1}{c}{HM}      & \multicolumn{1}{c|}{Acc.}  & \multicolumn{1}{c}{HM} & \multicolumn{1}{c|}{Acc.}  & \multicolumn{1}{c}{HM}      & \multicolumn{1}{c|}{Acc.}  & \multicolumn{1}{c}{HM}      & \multicolumn{1}{c}{Acc.}   \\ \midrule 
    CLIP          & 68.75 $\pm$ 0.00           & 65.39 $\pm$ 0.00           & 72.74 $\pm$ 0.00           & 67.28 $\pm$ 0.00           & 90.18 $\pm$ 0.00           & 85.40 $\pm$ 0.00           & 30.25 $\pm$ 0.00           & 23.94 $\pm$ 0.00           \\ 
    CoOp           & 68.22 $\pm$ 0.49           & 63.81 $\pm$ 0.44           & 78.33 $\pm$ 2.26           & 72.11 $\pm$ 2.36           & 86.65 $\pm$ 1.38           & 80.84 $\pm$ 1.50           & 29.38 $\pm$ 1.78           & 24.80 $\pm$ 1.23           \\ 
   CoCoOp        & 71.49 $\pm$ 0.62           & 67.75 $\pm$ 0.68           & 80.04 $\pm$ 1.46           & 71.95 $\pm$ 1.24           & 90.41 $\pm$ 0.24           & 85.61 $\pm$ 0.43           & 27.87 $\pm$ 11.36          & 21.46 $\pm$ 7.42           \\ 
    MaPLe           & 73.57 $\pm$ 0.77           & 69.97 $\pm$ 0.87         
    & 82.78 $\pm$ 0.69           & 77.32 $\pm$ 0.89           
    & \textbf{91.46 $\pm$ 0.14}  & \textbf{87.10 $\pm$ 0.22}           
    & 35.29 $\pm$ 0.58           & 27.63 $\pm$ 1.10           \\ \rowcolor{gray!20}
    BMIP     & \textbf{74.48 $\pm$ 0.85}  & \textbf{71.03 $\pm$ 0.68}  
    & \textbf{83.86 $\pm$ 1.70}  & \textbf{78.90 $\pm$ 1.57 } 
    & 90.86 $\pm$ 0.13  & 86.43 $\pm$ 0.30
    & \textbf{37.25 $\pm$ 0.93}  & \textbf{29.93 $\pm$ 2.75}  \\ 
    \midrule
    \midrule
    & \multicolumn{2}{c|}{SUN397}                             & \multicolumn{2}{c|}{DTD}                                & \multicolumn{2}{c|}{EuroSAT}                            & \multicolumn{2}{c}{UCF101}                              \\
    & \multicolumn{1}{c}{HM}      & \multicolumn{1}{c|}{Acc.}  & \multicolumn{1}{c}{HM}      & \multicolumn{1}{c|}{Acc.}  & \multicolumn{1}{c}{HM}      & \multicolumn{1}{c|}{Acc.}  & \multicolumn{1}{c}{HM}      & \multicolumn{1}{c}{Acc.}   \\ \midrule 
    CLIP           & 72.26 $\pm$ 0.00           & 62.57 $\pm$ 0.00           & 57.32 $\pm$ 0.00           & 44.56 $\pm$ 0.00           & 58.16 $\pm$ 0.00           & 41.40 $\pm$ 0.00           & 71.00 $\pm$ 0.00           & 64.97 $\pm$ 0.00           \\  
   CoOp           & 71.37 $\pm$ 1.21           & 61.82 $\pm$ 1.11           & 57.22 $\pm$ 2.37           & 48.18 $\pm$ 1.78           & 74.33 $\pm$ 4.35           & 59.65 $\pm$ 5.07           & 71.68 $\pm$ 2.84           & 65.41 $\pm$ 2.18           \\ 
    CoCoOp         & 77.17 $\pm$ 0.27           & 68.17 $\pm$ 0.33           & 60.59 $\pm$ 1.51           & 47.90 $\pm$ 1.43           & 73.77 $\pm$ 3.58           & 58.08 $\pm$ 1.49           & 76.59 $\pm$ 0.79           & 70.39 $\pm$ 1.25           \\ 
    MaPLe          & \textbf{ 79.58 $\pm$ 0.13 } & \textbf{70.90 $\pm$ 0.22}   
    & 64.49 $\pm$ 3.73           & 54.13 $\pm$ 2.19       
    & 81.43 $\pm$ 0.53           & 70.33 $\pm$ 2.74          
    & 80.69 $\pm$ 0.26           & 73.53 $\pm$ 0.45           \\ \rowcolor{gray!20}
    BMIP     & 79.02 $\pm$ 0.24  & 70.57 $\pm$ 0.47
    & \textbf{67.02 $\pm$ 0.90}  & \textbf{54.90 $\pm$ 1.28}  
    & \textbf{86.10 $\pm$ 1.58}  & \textbf{73.77 $\pm$ 1.10}  
    & \textbf{82.29 $\pm$ 0.96}  & \textbf{75.00 $\pm$ 0.80}  \\ 
    \bottomrule
    \end{tabular}
    \caption{Performance comparison on 11 datasets using ViT-B/16 architecture. The best performance is in bold.}
    \label{tab:full-exps-vit16}
\end{table*}

\begin{table*}[t]
  \centering
  \setlength{\tabcolsep}{1mm}
  \begin{tabular}{lcccccccccccc}
    \toprule
    ~ & \textbf{Source} & \multicolumn{10}{c}{\textbf{Target}} & \multirow{2}{*}{\textbf{Average}} \\ 
    \cmidrule(r){2-2} \cmidrule(r){3-12}
~ & ImageNet & Caltech & Pets & Cars & Flowers & Food & Aircraft & SUN & DTD & EuroSAT & UCF &  \\
    \midrule
    CLIP   & 66.70 & 93.30 & 89.10 & 65.70 & 70.70 & 85.90 & 24.80 & 62.60 & 44.30 & 48.30 & 67.60 & 65.24 \\
    CoOp   & \textbf{71.51} & 93.70 & 89.14 & 64.51 & 68.71 & 85.30 & 18.47 & 64.15 & 41.92 & 46.39 & 66.55 & 63.88 \\
    CoCoOp & 71.02 & \textbf{94.43} & 90.14 & 65.32 & 71.88 & 86.06 & 22.94 & 67.36 & 45.73 & 45.37 & 68.21 & 65.74 \\
    MaPLe  & 70.72 & 93.53 & \textbf{90.49} & 65.57 & \textbf{72.23} & 86.20 & \textbf{24.74} & 67.01 & \textbf{46.49} & 48.06 & 68.69 & 66.30 \\ \midrule
    \rowcolor{gray!20}
    BMIP  & 70.86 & 94.13 & 90.13 & \textbf{66.03} & 72.13 & \textbf{86.23} & 24.10 & \textbf{67.30} & 45.97 & \textbf{49.63} & \textbf{68.96} & \textbf{66.86} \\
    \bottomrule
    \end{tabular}
    \caption{Comparison of BMIP with existing approaches on cross-dataset transfer task.}
    \label{tab:cross-dataset}
\end{table*}

\subsection{Open-World Generalization Task}
To assess the open-world robustness of the BMIP approach, the average performance across all datasets, as well as the detailed performance on each dataset measured by two metrics, i.e., the harmonic mean (HM) of the accuracies for the base and new classes and Accuracy, are reported. 
The average performance and variance under three runs are presented in \Cref{tab:full-exps-vit16}. Compared to the SOTA method MaPLe, BMIP demonstrates performance improvements on both base and novel classes across most datasets. In the remaining three datasets, the BMIP method achieves competitive performance compared to the MaPLe method and delivers the best average performance on both the HM and Accuracy metrics.
BMIP shows an absolute average performance gain of 0.82\% over MaPLe, when considering both base and novel classes simultaneously. 
Additionally, BMIP performs exceptionally well on datasets with imbalanced text and image information~\cite{nilsback2008automated,helber2019eurosat,zang2022unified}, such as those with high intra-class visual variance (e.g., EuroSAT, from 81.43\% to 86.10\%) and low inter-class text variance (e.g., Flowers102, from 82.78\% to 83.86\%), reflecting our motivation that BMIP can overcome the shortcomings of single-modal prompt learning methods.
The above results highlight the crucial role of the independent prompts of each modality and the strong alignment between vision and language in enhancing generalization capacity. 

\subsection{Cross-Dataset Transfer Task}
To evaluate the transfer ability of our method, we finetune multi-modal prompts using the ImageNet dataset and directly transfer them to the other 10 datasets. 
\Cref{tab:cross-dataset} presents a performance comparison between CoOp, CoCoOp, MaPLe, and BMIP.
BMIP exhibits competitive performance across 10 datasets, achieving the highest average accuracy of 66.86\%, with a particularly notable accuracy of 49.63\% on EuroSAT. Although CoOp performs the best on the source dataset, its performance degrades on other datasets.
These results suggest that using bi-directional modality interaction within BMIP aids in better cross-dataset generalization, exceeding MaPLe, highlighting the significance of incorporating bi-directional information in prompt learning.

\subsection{Domain Generalization Task}
We assess the robustness of the mentioned methods trained on ImageNet to various out-of-distribution (OOD) datasets. BMIP outperforms MaPLe on three of four out-of-distribution datasets, enhancing the OOD average classification accuracy to 60.40\% and achieving an overall average accuracy of 62.50\% on both training and testing dataset, as shown in \Cref{tab:domain-generalization}.
The results demonstrate that BMIP effectively addresses open-world challenges, including open-world generalization and domain generalization, highlighting the model’s strong performance in these areas.

\begin{table*}[t]
\centering
\begin{tabular}{@{}cccccccc@{}}
\toprule
 \multirow{2}{*}{Method} & \textbf{Source} & \multicolumn{4}{c}{\textbf{Target}} & \multirow{2}{*}{\textbf{Average}} & \multirow{2}{*}{\begin{tabular}{@{}c@{}}\textbf{OOD}\\\textbf{Average}\end{tabular}} \\
 \cmidrule(r){2-2} \cmidrule(r){3-6}
 & ImageNet & ImagenetV2 & ImageNet-S & ImageNet-A & ImageNet-R  \\ \midrule
CLIP & 66.73 & 60.83 & 46.15 & 47.77 & 73.86 & 59.07 & 57.15 \\
CoOp & \textbf{71.51} & 64.20 & 47.99 & 49.71 & 75.21 & 61.72 & 59.28 \\
CoCoOp & 71.02 & 64.07 & 48.75 & 50.63 & 76.18 & 62.13 & 59.91 \\
MaPLe & 70.72 & 64.07 & \textbf{49.15} & 50.90 & 76.98 & 62.36 & 60.28 \\
\midrule
\rowcolor{gray!20}
BMIP & 70.86 & \textbf{64.23} & 49.13& \textbf{51.06} & \textbf{77.20} & \textbf{62.50} & \textbf{60.40} \\ \bottomrule
\end{tabular}
\caption{Comparison of BMIP with existing approaches in domain generalization setting.}
\label{tab:domain-generalization}
\end{table*}

\subsection{Flexibility of BMIP} 
BMIP focuses on prompt learning for modality interaction and can be becombined with any prompt-based model, using MaPLe as the foundational model. Therefore, to verify the flexibility of BMIP, we combine BMIP with two representative methods, PromptSRC~\cite{promptsrc} and CoPrompt~\cite{Coprompt}, which represent the latest approaches utilizing additional knowledge, such as regularized information, to enhance model training. \Cref{com-exp} indicates that BMIP brings a significant performance improvement to these methods, demonstrating BMIP’s superior capability as a foundational model compared to MaPLe.

\begin{table}[t]
    \centering
    \begin{tabular}{ccccc}
    \toprule
        Method & HM & Acc.  \\ 
        \midrule
        CLIP & 71.70 $\pm$ 0.00 & 63.92 $\pm$ 0.00 \\ \hline
        MaPLe & 78.22 $\pm$ 0.26 & 71.76 $\pm$ 0.28 \\
        \rowcolor{gray!20}
        +BMIP & \textbf{79.03 $\pm$ 0.13} & \textbf{72.54 $\pm$ 0.26} \\ \hline
        PromptSRC & 79.67 $\pm$ 0.46  & 73.43 $\pm$ 0.35 \\
        \rowcolor{gray!20}
        +BMIP & \textbf{80.03 $\pm$ 0.22} & \textbf{73.97 $\pm$ 0.43} \\ \hline
        CoPrompt & 78.99 $\pm$ 0.25 & 71.48 $\pm$ 0.73 \\
        \rowcolor{gray!20}
        +BMIP & \textbf{79.54 $\pm$ 0.28} & \textbf{72.35 $\pm$ 0.32} \\
        \bottomrule
    \end{tabular}
    \caption{Performance improvements from combining three baselines with BMIP (average across 11 datasets)}
    \label{com-exp}
\end{table}

\subsection{Ablation Studies}
In our ablation studies, we explore various aggregation functions to determine the individual contributions of the BMIP components to their overall performance, thus validating the intuition derived from our corollary. To ensure that the performance improvement of BMIP is not attributed to the number of parameters, we conduct comparative experiments that equalize the number of parameters between MaPLe and BMIP.

\textbf{Compare with Other Aggregation Functions.} We describe and compare different aggregation functions in more detail in Appendices B and C to verify the benefits of the learnable weighted aggregation function over direct prompt modification functions, such as Addition, Attention, and Joint. As we can see from~\cref{Agg-fun}, the comparison with Addition highlights the importance of our learnable weights, while the contrast with Attention demonstrates the reliability of learnable weights over similarity-based prompt calculations. Furthermore, the comparison with concatenation-based prompt aggregation underscores the effectiveness of our replacement strategy in reducing prompt redundancy. The comparisons above validate the intuition presented in our analysis: the prompts in VLMs can be mutually replaced and optimized simultaneously and BMIP improves the trainability of prompts. We provide a detailed report on the performance of different aggregation functions across all settings in Appendix C.

\begin{table}[ht]
    \centering
    \setlength{\tabcolsep}{1mm}
    \begin{tabular}{ccc}
    \toprule
    \multirow{2}{*}{\parbox{2cm}{\centering \textbf{Aggregation \\ Method}}} & \multicolumn{2}{c}{\textbf{Open-World Generalization}} \\ 
        ~ & \parbox{2cm}{\centering HM} & \parbox{1.7cm}{\centering Acc.}  \\ 
        \midrule
        IVLP & 77.51 & 71.18 \\
        CoCoOp & 75.83 & 67.67 \\
        MaPLe & 78.18 & 71.72 \\
        $\text{MaPLe}^\dagger$ & 77.19 & 71.66 \\
        \midrule
        Addition & 78.40 & 71.42 \\ 
        Attention & 77.74 & 71.56 \\
        Joint & 78.79 & 71.72 \\ \midrule \rowcolor{gray!20}
        BMIP & \textbf{79.04} & \textbf{72.17} \\
    \bottomrule
    \end{tabular}
    \caption{Comparison of BMIP with different methods in open-world generalization tasks. IVLP represents independently trained vision and language prompts.}
    \label{Agg-fun}
\end{table}

\textbf{Number of Parameters.} 
To ascertain that BMIP's enhanced performance is not due to an increase in parameter count relative to MaPLe, we equalize the number of parameters between MaPLe (denoted \(\text{MaPLe}^\dagger\)) and BMIP. \Cref{Agg-fun} demonstrates that \(\text{MaPLe}^\dagger\) tends to overfit to base classes when parameters are increased, whereas BMIP maintains robust performance despite a higher parameter count. This indicates the efficacy of BMIP's bidirectional modality interaction framework in harmonizing prompt information across modalities. We also assess the impact of varying prompt depths and lengths in Appendices D and E, which proves our intuition about prompt depth.
% \begin{table}[ht]
%     \centering
%     \begin{tabular}{ccccc}
%     \toprule
%         Method & Base & Novel & HM & Acc.  \\ 
%         \midrule
%         CLIP & 69.34 & 74.22 & 71.70 & 63.92 \\
%         CoOp & 82.69 & 63.22 & 71.66 & 65.57 \\ 
%         MaPLe & 81.96 & 74.73 & 78.18 & 71.72 \\ 
%         $\text{MaPLe}^\dagger$ & 80.23 & 74.37 & 77.19 & 71.66 \\
%         \midrule
%         BMIP & \textbf{82.97} & \textbf{75.47} & \textbf{79.04} & \textbf{72.17} \\
%         \bottomrule
%     \end{tabular}
%     \caption{Performance of MaPLe and BMIP using the same number of parameters.}
%     \label{ab-exp}
% \end{table}
% UPT & 81.37 & 73.66 & 77.32 & 70.63 \\

\section{Conclusion}
Prompt learning is regarded as a pivotal technique for adapting pre-trained VLMs to specific downstream tasks. However, existing methods mainly focus on either single-modal prompts or uni-directional modality interaction, which neglects the potent alignment effects that arise from the modalities interaction, causing performance degradation on various datasets. This paper proposes a novel bi-directional modality interaction prompt learning method (BMIP) that includes advanced mechanisms for deep language prompt learning, deep vision prompt learning, and most notably, the interaction between vision and language modalities. Furthermore, we propose a new evaluation paradigm termed \emph{open-world generalization} which offers a more realistic evaluation of prompt learning. Extensive experimental results demonstrate the effectiveness of the proposed BMIP method across various evaluation tasks, particularly in handling datasets with unbalanced text and image variances, such as EuroSAT, and the proposal is flexible enough to be applied to further improve the performance of other methods consistently.

\bibliographystyle{named}
\bibliography{ijcai25}

\end{document}